\pdfoutput=1
\documentclass[a4paper,twoside]{article}

\usepackage{epsfig}
\usepackage{subcaption}
\usepackage{calc}
\usepackage{amssymb}
\usepackage{amstext}
\usepackage{amsmath}
\usepackage{amsthm}
\usepackage{multicol}
\usepackage{pslatex}
\usepackage{apalike}
\usepackage{algorithm2e}
\usepackage[bottom]{footmisc}
\usepackage[colorlinks=true,
            linkcolor=blue,
            filecolor=magenta,
            urlcolor=cyan,
            citecolor=blue]{hyperref}
\usepackage{enumitem}
\usepackage{SCITEPRESS}     % Please add other packages that you may need BEFORE the SCITEPRESS.sty package.

\newcommand{\sect}{§~}
\newcommand{\sectref}[1]{\hyperref[#1]{\sect\ref*{#1}}}

\newtheorem{conjecture}{Conjecture}

\begin{document}

\title{Agent-Centric Projection of Prompting Techniques and Implications for Synthetic Training Data for Large Language Models}

\author{\authorname{Dhruv Dhamani\orcidAuthor{0009-0003-8226-7621} and Mary Lou Maher\orcidAuthor{0000-0002-4150-0322}}
\affiliation{\sup{1}University of North Carolina, Charlotte}
\email{ddhamani@charlotte.edu, m.maher@charlotte.edu}
}

\keywords{
  Large Language Models (LLMs),
  Task-oriented LLM System,
  Prompt Engineering,
  Large Language Model-based Agent,
  LLM-based Multi-agent System,
  Synthetic Training Data,
  Artificial Intelligence in Problem Solving
}

\abstract{Recent advances in prompting techniques and multi-agent systems for Large Language Models (LLMs) have produced increasingly complex approaches. However, we lack a framework for characterizing and comparing prompting techniques or understanding their relationship to multi-agent LLM systems. This position paper introduces and explains the concepts of linear contexts (a single, continuous sequence of interactions) and non-linear contexts (branching or multi-path) in LLM systems. These concepts enable the development of an agent-centric projection of prompting techniques, a framework that can reveal deep connections between prompting strategies and multi-agent systems. We propose three conjectures based on this framework: (1) results from non-linear prompting techniques can predict outcomes in equivalent multi-agent systems, (2) multi-agent system architectures can be replicated through single-LLM prompting techniques that simulate equivalent interaction patterns, and (3) these equivalences suggest novel approaches for generating synthetic training data. We argue that this perspective enables systematic cross-pollination of research findings between prompting and multi-agent domains, while providing new directions for improving both the design and training of future LLM systems.}

\onecolumn \maketitle \normalsize \setcounter{footnote}{0} \vfill

\section{\uppercase{Introduction}}\label{sec:introduction}

Large Language Models (LLMs) are a recent development in Generative Artificial Intelligence that can mimic human-like behavior~\cite{parkGenerativeAgentsInteractive2023}, especially in conversations~\cite{caiDoesChatGPTResemble2023}. LLMs have also shown a kind of general intelligence~\cite{radford_language_2019,yogatamaLearningEvaluatingGeneral2019}. Central to harnessing the capabilities of LLMs is the concept of prompting, a strategy that significantly influences task performance by instructing LLMs in specific ways~\cite{chenUnleashingPotentialPrompt2023}.

\textsc{Hypothesis and Goals:} In this position paper, we hypothesize that viewing prompting techniques through a proposed agent-centric lens can help uncover structural equivalences between single-LLM prompting and multi-agent approaches. Our goal is to (1) introduce a unified framework for comparing these techniques, (2) develop and examine conjectures about their relationship, and (3) outline how this perspective can inform the generation of synthetic training data.

Consider a simple math problem. When we directly prompt an LLM, ``What is 13 $\times$ 27?'', we might receive a single numeric answer. However, when we ask, ``Let’s solve this step by step: what is 13 $\times$ 27?'', we explicitly prompt for intermediate reasoning plus the final result~\cite{kojima_large_2023,yuBetterChainofThoughtPrompting2023}. While both prompts seek the same final answer, are they both still the same problem if one has a different ``correct'' answer?

Another approach to improving end-task performance when using LLMs has been to incorporate ``reasoning''~\cite{openai_learning_2024}. The model outputs arbitrarily long ``reasoning traces'' before responding to the prompts. These traces are sequences of natural language statements like ``Let's first understand the input and output formats''. OpenAI o1 is a single large language model or agent.

If we simply added role identifiers before each statement --- ``Analyst: Let's first understand the input and output formats'' --- would it suddenly qualify as a multi-agent system?

What about approaches where an LLM analyzes problems from multiple perspectives in separate conversations before merging all perspectives together in another conversation~\cite{saha_branch-solve-merge_2023}? Is each conversation a different ``agent'' performing a subtask? Is this a multi-agent system?

We argue that these questions can be systematically addressed by viewing prompting techniques through an agent-centric lens. By developing the concepts of linear and non-linear contexts in LLM systems, we shed light on possible connections between single-LLM prompting techniques and multi-agent systems. We discuss implications for the future of LLM systems, from enabling cross-pollination of research findings between prompting techniques and multi-agent systems to suggesting novel approaches for generating synthetic training data that could enhance capabilities in both domains.

To develop this argument, we first establish foundational definitions and examine previous work on prompting techniques and task-oriented LLM systems (\sectref{sec:definitions_and_prior_work}). Building on these foundations, we present our framework for agent-centric projection and explore its implications for both system design and training (\sectref{sec:projections_and_conjectures}). We conclude by discussing the larger impact of this perspective on future research in LLM systems (\sectref{sec:summary_and_conclusion}).

\section{\uppercase{Definitions and Prior Work}}\label{sec:definitions_and_prior_work}

A task-oriented LLM system is a Large Language Model (LLM) system configured to perform specific tasks, rather than open-ended conversations\footnote{In~\citenp{xi_rise_2023}, the authors describe task-oriented deployments of LLM-based agents, which we generalize to simply task-oriented LLM systems}. Such systems have shown promise in complex tasks such as software development~\cite{hong_metagpt_2023}, where the system must manage multiple rounds of interaction, maintain context throughout iterations, and often collaborate with other systems or agents to complete the task.

We begin by defining a minimal task-oriented LLM system (\sectref{sec:minimal_llm_system}), with particular attention to how such systems manage context across multiple interactions. We then examine prompting techniques (\sectref{sec:prompting_techniques}), focusing on how different approaches to prompting lead to different patterns of context creation and management. These patterns form the basis for our novel concepts of linear and non-linear contexts, which enable an agent-centric projection of prompting techniques.

\subsection{Minimal Task-oriented LLM System}\label{sec:minimal_llm_system}

A minimal task-oriented LLM system is a minimal LLM system that can be instructed to solve tasks. Thus, we start by defining a minimal LLM system.

Large Language Models are auto-regressive models that accept input tokens and use them as history (often referred to as context), to compute probabilities of all tokens in their vocabulary as the next token. We can sample from this probability distribution using a sampling/decoding algorithm to generate text. This process is then repeated until the LLM predicts a special token, or a special sequence of tokens, that marks the end of the text~\cite{feuerriegel_generative_2023}\footnote{The Generative AI system description in~\citenp{feuerriegel_generative_2023} includes any UI components as part of the Generative AI system, and we use a modified definition that only includes the language model and sampling/decoding procedure here.}.

\begin{figure}[!ht]
  \centering
  \includegraphics[width=0.96\columnwidth]{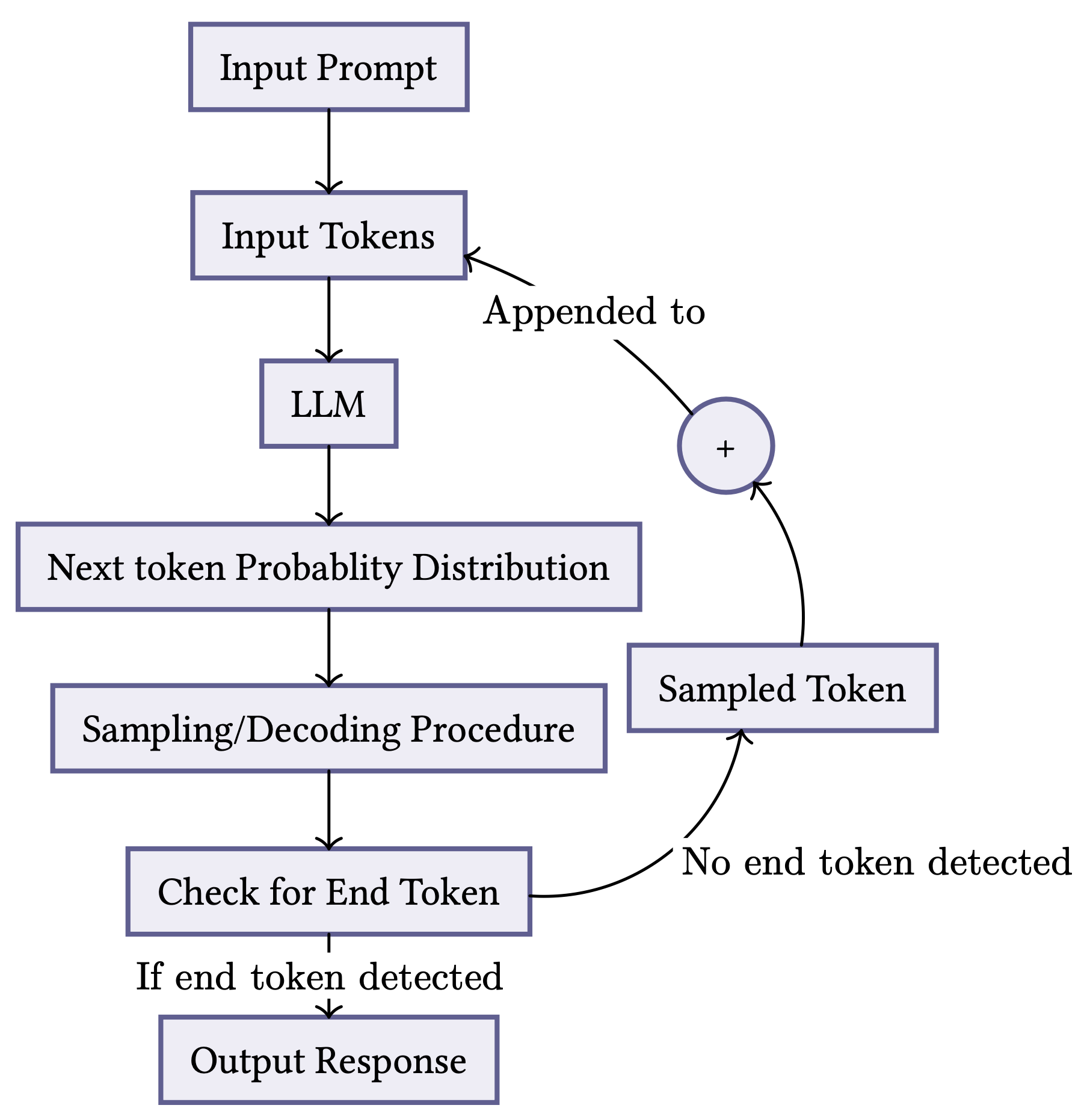}
  \caption{A bare-bones LLM system.}\label{fig:barebones_llm}
\end{figure}

We call this a bare-bones LLM system (see Figure~\ref{fig:barebones_llm}) as it contains the minimal components needed for text generation, without additional components to help with context management. Every time an LLM is prompted with context $C_n$, it generates a response $R_n$ that would need to be stored in context $C_{n+1}$ for the next prompt, assuming multiple rounds of instruction and response generation are required.

For systems oriented towards solving even moderately complex tasks, context management becomes quickly cumbersome. For example, in~\citenp{saha_branch-solve-merge_2023}, the authors describe a system in which $m$ branches are created from a LLM response to a prompt $C_n$, producing a set of $m$ responses $r = \{R_{n_1}, R_{n_2}, \ldots, R_{n_m}\}$. Then, they take all of $r$, and transform it into a prompt $C_{n+1}$, in which they instruct the LLM to merge all responses in $r$ into a single response $R_{n+1}$, which potentially needs to be stored in context $C_{n+2}$ for the next prompt. A similarly complex system is described in~\citenp{ningSkeletonofThoughtLargeLanguage2023}, and we will examine more examples in our discussion of prompting techniques.

If we define a minimal LLM system without describing how context is managed, it would be too difficult to compare different systems and apply learnings from one researched system to another. Because of this, we include a description of a minimal context management subsystem within our definition of a minimal task-oriented LLM system.

Specifically, we include a context store $CS$. Initially, the context store $CS$ is empty and the first time an LLM is provided with context $C_1$ to generate $R_1$, both the prompt and the response are permanently appended to $CS$. For all future requests, the LLM is first provided with a sliding window of content from $CS$ as context, to which it appends the prompt $C_n$ to generate $R_n$. Once the response $R_n$ is generated, both the prompt and the response are permanently appended to $CS$. The model then closely matches messaging: the context store $CS$ acts as chat history, users send messages, and the LLM responds.

\begin{figure}[!ht]
  \centering
  \includegraphics[width=0.78\columnwidth]{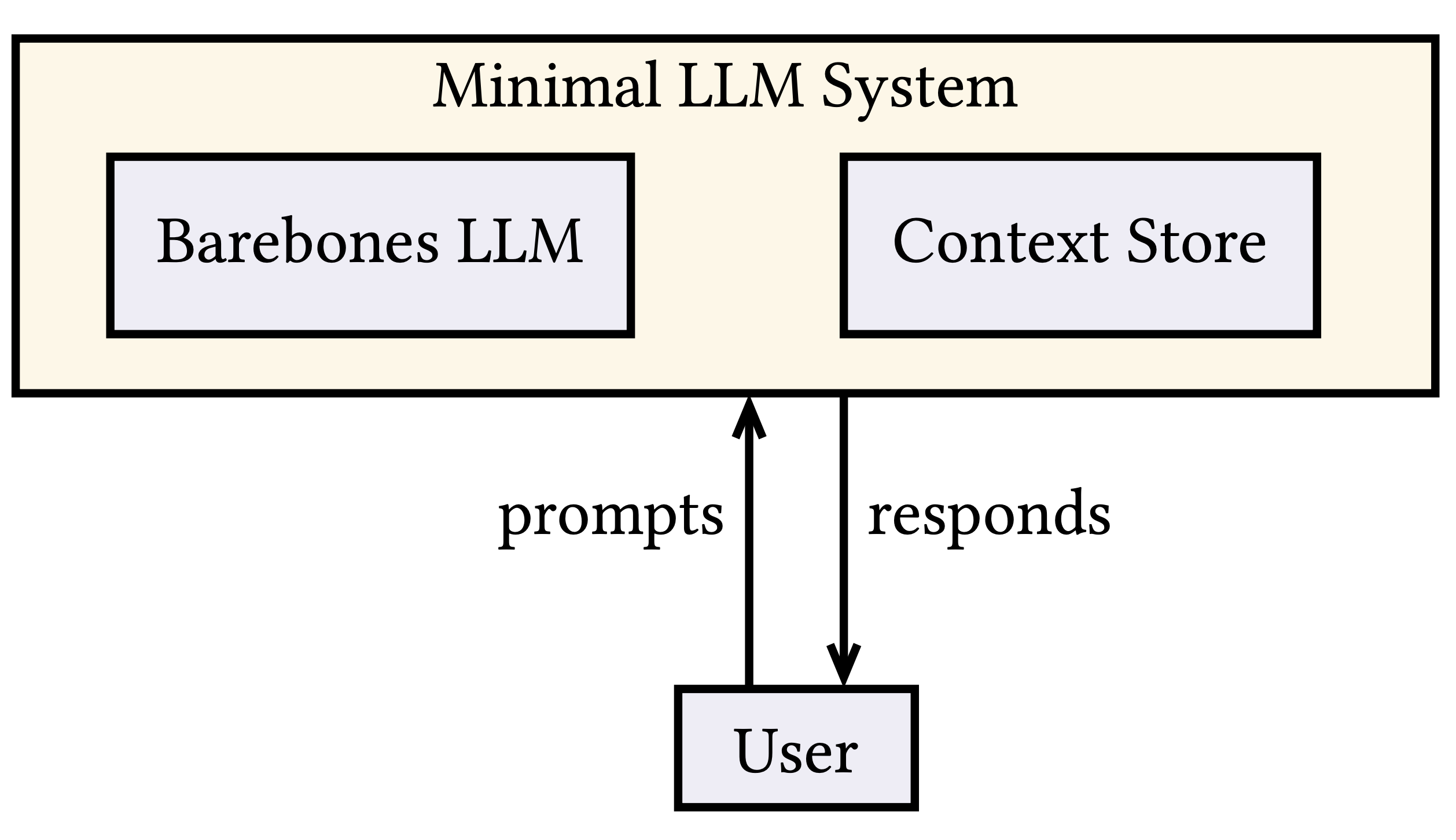}
  \caption{A minimal LLM system that includes a context store.}\label{fig:minimal_llm_system}
\end{figure}

We discuss the implications of this composition of a minimal LLM system in \sectref{sec:agent_centric_projection}.

\subsection{Prompting Techniques}\label{sec:prompting_techniques}

Prompting refers to the act of constructing and providing input text (a prompt) to an LLM. In the context of task-oriented LLM systems, prompt engineering can be defined as iteratively creating and adapting a prompt for a given LLM and task pair.

The way an LLM is prompted significantly affects task performance~\cite{nori_can_2023,savage_diagnostic_2023}. There are many surprising results in this area, such as letting an LLM know that solving a task ``is very important to my career'' can improve task performance~\cite{li_large_2023}.

Such results can be explained by research such as~\citenp{hendel_-context_2023}, which shows that in-context learning creates task vectors or representations within the LLM that increase the probability of correct task completion. Other research has shown that it is possible to ``search'' for prompts that are more likely to lead to success, analogous to finding task vectors that are more likely to lead to success. In~\citenp{zou_universal_2023}, the authors were able to procedurally find adversarial prefixes, which, when added to prompts, result in LLMs breaking their alignment and engaging in unsafe behavior.

All of these are examples of modifying the prompt without changing the actual task/problem definition, to make the successful completion of the intended task more likely. However, researchers are prone to modifying the prompts in a manner that changes the task, rather than modifying the prompts in a manner that improves task performance.

For example, when using Chain-of-Thought prompting~\cite{wei_chain--thought_2023}\footnote{As described in the paper, one is to also provide in-context examples, but this is unnecessary, as shown in~\citenp{kojima_large_2023}}, or when asking LLMs to think step-by-step~\cite{kojima_large_2023} -- the task meaningfully changes. It goes from instructing LLMs to give me an answer now to asking it to first plan out a solution, and then share an answer. This is a different task being solved, even though the final deliverable (the answer) is the same. It should be a given that LLMs have different capabilities for different tasks.

This is not to say that we shouldn't instead solve equivalent tasks that LLMs are more suitable to, but that it is problematic to have prompt modification (that leaves instructions/task definition intact) to instruction modification in the same category. Thus, we make the distinction between prompt engineering and instruction engineering:

\begin{itemize}
\item \textsc{Prompt Engineering}: The act of modifying the prompt without changing the actual task/problem definition or adding relevant knowledge/information, to make the successful completion of the intended task more likely. We restrict the addition of relevant knowledge/information to LLM augmentation to avoid an overlap.

\item \textsc{Instruction Engineering}: The act of modifying the prompt in a manner that changes the task/problem to an equivalent task/problem that the LLMs are more suitable for, such that the final deliverable (the answer) is the same.
\end{itemize}

In~\citenp{bestaGraphThoughtsSolving2023a}, the authors describe a taxonomy of techniques to improve reliability in task-oriented text generation:

\begin{itemize}
\item \textsc{Input-Output}: The LLM is directly being instructed to respond with the result of a prompted task.

\item \textsc{Input-Output with additional steps}: The LLM is instructed to perform additional steps before or after generating a result for a prompted task, such as reflecting on its response and refining it, or creating a plan~\cite{madaanSelfRefineIterativeRefinement2023,wei_chain--thought_2023}.

\item \textsc{Single Input-Many Output}:\footnote{Referred to as Multiple CoTs in~\citenp{bestaGraphThoughtsSolving2023a}} The LLM is passed the same input prompt multiple times, and elaborate mechanisms are used to choose the final answer~\cite{wangSelfConsistencyImprovesChain2022}.

\item \textsc{Input with Non-linear intermediary steps}:\footnote{This is not described in~\citenp{bestaGraphThoughtsSolving2023a}} LLM branches into multiple paths (through variations of an input prompt), generating multiple responses as additional steps, and then merges them into a single response~\cite{saha_branch-solve-merge_2023,ningSkeletonofThoughtLargeLanguage2023}.

\item \textsc{Tree of Thoughts}: An elaborate method described in~\citenp{yaoTreeThoughtsDeliberate2023}, where many intermediate thought branches are explored, backtracked, and pruned until a final answer is settled on.

\item \textsc{Graph of Thoughts}: An elaborate method described in~\citenp{bestaGraphThoughtsSolving2023a}, where intermediate thoughts are modeled as a connected graph, and the LLM traverses the graph to settle on a final answer.
\end{itemize}

The way these techniques manage context differs significantly, from linear interactions to branching paths of thought. In the next section, we introduce the concepts of Linear and Non-Linear contexts to formalize these differences, and show how this formalization enables an Agent-centric projection of prompting techniques with potential implications for synthetic training data generation.

\section{\uppercase{Framework and Conjectures}}\label{sec:projections_and_conjectures}

In research and practice, LLM systems exhibit different patterns in how they manage context and generate responses. We argue that these patterns can be understood through a theoretical framework that connects prompting techniques with multi-agent systems, revealing opportunities for improving both system design and training. In this section, we first introduce a formal categorization of context management patterns in LLM systems (\sectref{sec:linear_nonlinear_context}). Building on this foundation, we develop an agent-centric projection of prompting techniques (\sectref{sec:agent_centric_projection}) that reveals deep connections between seemingly disparate areas. Finally, we explore how this unified perspective suggests novel approaches to synthetic training data generation (\sectref{sec:synthetic_training_data}), with potentially far-reaching implications for improving LLM capabilities.

\subsection{Linear and Non-linear Context in LLM Systems}\label{sec:linear_nonlinear_context}

To formally characterize how task-oriented LLM systems manage context and generate responses, we develop a framework based on message flow patterns. Building on the minimal task-oriented LLM system concept (\sectref{sec:minimal_llm_system}), we analyze how the context store maintains sequences of messages $M = \{(C_n, R_n)\}_{n=1}^N$, where each response $R_n$ is generated using all previous context-response pairs.

Using this foundation, we propose a method for classifying prompting techniques and their resulting task-oriented LLM systems into two categories based on their context management patterns.

\textsc{Prompting techniques with Linear context} --- where there exists exactly one continuous sequence of messages $M = \{(C_n, R_n)\}_{n=1}^N$ that contains all generated messages and input contexts in the correct chronological order.

All Input-Output and Input-Output with additional steps techniques (as described in \sectref{sec:prompting_techniques}) can be classified as having a linear context, as they all involve a single continuous sequence of messages.

For example, consider Self-Refine~\citenp{madaanSelfRefineIterativeRefinement2023}, where each response is iteratively refined using all previous context-response pairs until a stop condition is met.

\textsc{Prompting techniques with Non-linear context} -- where there cannot always be one continuous sequence of messages that contains all input context and generated messages in the correct chronological order. Instead, there can be multiple branches of conversation possible, each with its own continuous sequence of messages $\{M_1, M_2, \ldots, M_n\}$.

All single input-many output, input with non-linear intermediary steps, tree of thought, and graph of thought techniques, (as described in \sectref{sec:prompting_techniques}) can be classified as having a non-linear context, as they all potentially involve sequences of conversation $M$.
 
\begin{figure}[!ht]
 \centering
 \includegraphics[width=0.60\columnwidth]{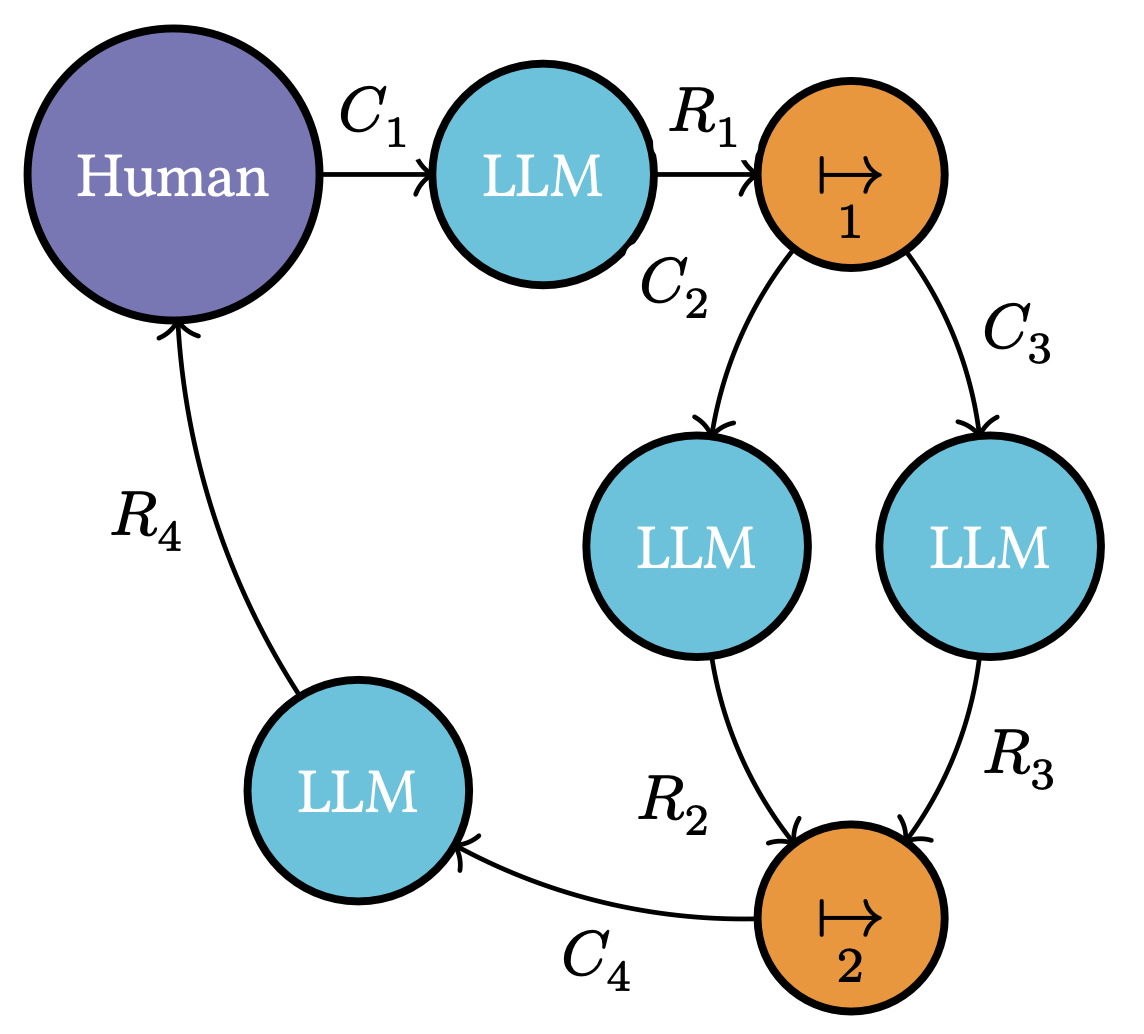}
 \caption{An example of a prompting technique with non-linear context.}\label{fig:non_linear_prompting}
\end{figure}

For example, consider a simplified version of BRANCH-SOLVE-MERGE, first described in~\citenp{saha_branch-solve-merge_2023}, and as visualized in Figure~\ref{fig:non_linear_prompting}. The figure depicts a task-oriented LLM system that helps the user (the human) make decisions. First, the human first instructs the system to make a decision. The system uses the instructions to create an input context for an LLM (context $C_1$) and uses it to generate a response $R_1$. $R_1$ is then used to create two new prompts (via an algorithmic transformation depicted in the figure as \raisebox{2.4pt}{$\underset{1}{\mapsto}$}), one in which the LLM is tasked with reflecting on the drawbacks of this decision (in context $C_2$) and another where the LLM is tasked with reflecting on the benefits of this decision (in context $C_3$). Finally, another prompt is created where both reflections (responses $R_2$ and $R_3$) are considered (using another algorithmic transformation \raisebox{2.4pt}{$\underset{2}{\mapsto}$}) to create a new prompt (context $C_4$) which is used to generate a final decision within response $R_4$. $R_4$ is then reported to the user as the final decision.

As long as any task-oriented system is using LLMs, it will always have one or more continuous streams of messages $M$ as described. This means that all task-oriented LLM systems and all prompting techniques can be classified as having either linear or non-linear contexts.

This fundamental dichotomy between linear and non-linear contexts provides a powerful lens through which to analyze LLM systems. As we will show in the next section, it reveals surprising connections between prompting techniques and multi-agent systems that can inform both system design and training approaches.

\subsection{Agent-Centric Projection of Prompting Techniques}\label{sec:agent_centric_projection}

In the previous section (\sectref{sec:linear_nonlinear_context}), we classify all prompting techniques and all resulting task-oriented LLM systems they bore into either having a linear or non-linear context. This decision and the overall definition have the following implications:

\begin{itemize}
\item Research on techniques for reliable, task-oriented text generation that involve linear context can be modeled as a kind of two-agent system (the human instructing the LLM being the second agent, as we also see in~\citenp{xi_rise_2023,wu_autogen_2023}).

\item Research on techniques for reliable, task-oriented text generation that involves non-linear context can be modeled to be a kind of multi-agent system, where each ``branch'' of conversation $M$ can be considered to have occurred with a different agent.

\begin{figure}[!ht]
 \centering
 \includegraphics[width=0.60\columnwidth]{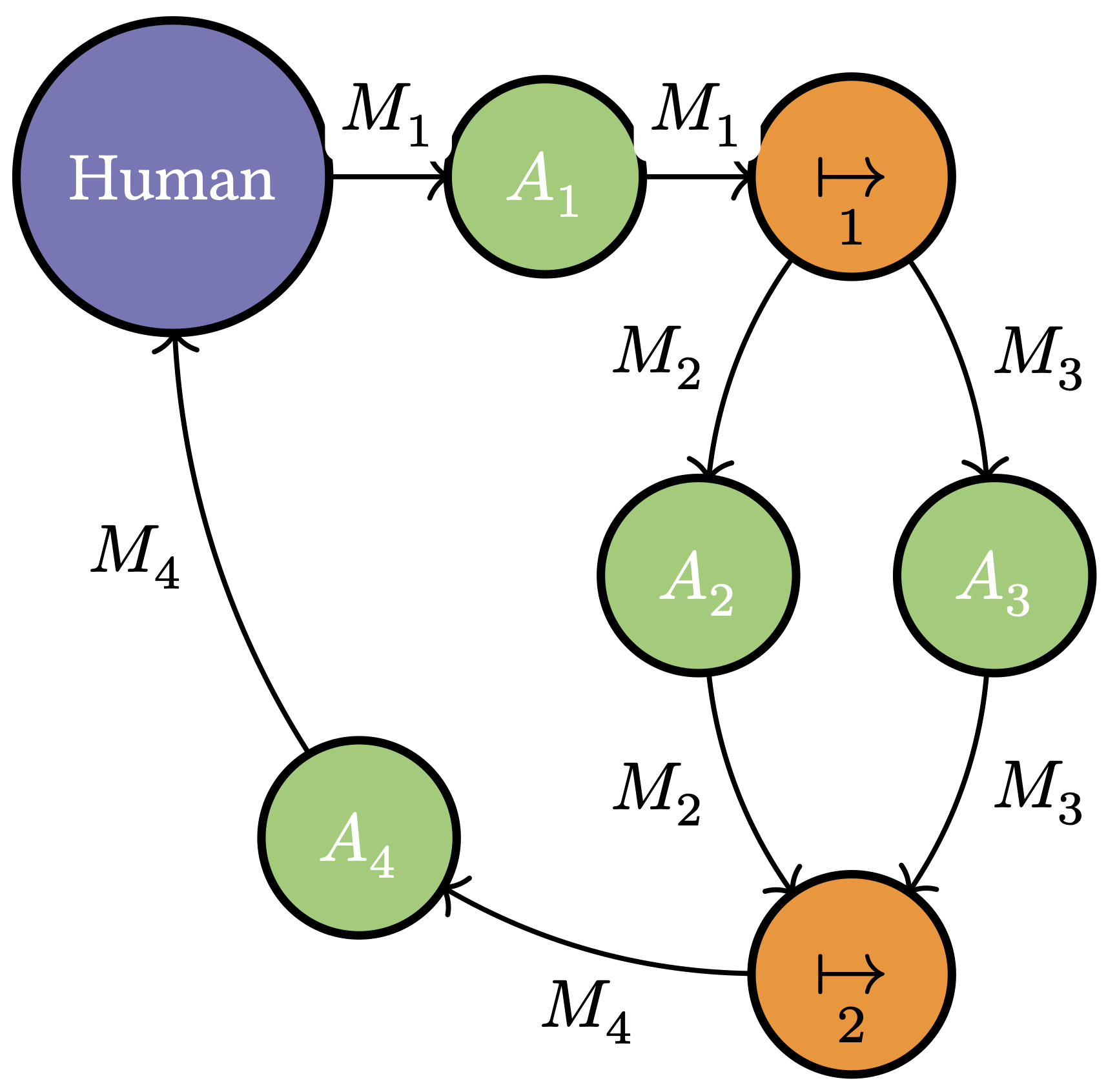}
 \caption{The prompting technique from Figure~\ref{fig:non_linear_prompting} is modeled as a multi-agent system.}\label{fig:non_linear_prompting_agent}
\end{figure}

 For example, In Figure~\ref{fig:non_linear_prompting_agent}, we show how the prompting technique from Figure~\ref{fig:non_linear_prompting} can be modeled as a multi-agent system. Each continuous linear sequence of messages $M_1 = \{C_1, R_1\}$, $M_2 = \{C_2, R_2\}$, $M_3 = \{C_3, R_3\}$, and $M_4 = \{C_4, R_4\}$ can be considered to have occurred with a different agent. Using this approach, we can model any prompting technique with non-linear context as a multi-agent system.

It would also help to note that each continuous sequence of messages $M_n$ in Figure~\ref{fig:non_linear_prompting_agent} is essentially a minimal task-oriented LLM system, as described in \sectref{sec:minimal_llm_system}. This means that we can substitute each such minimal system with a more complicated task-oriented system if needed.

In Figure~\ref{fig:agent_centric_prompting}, we show a more realistic example of a multi-agent system, designed to replicate the behavior of the prompting technique in Figure~\ref{fig:non_linear_prompting}. Here, the major changes are that the agents communicate with each other using tools, meaning all communication is bidirectional (say, if an agent wants to ask a clarifying question) and that the algorithmic transformations $\underset{1}{\mapsto}$ and $\underset{2}{\mapsto}$ are now present each as a tool available to Agents $A_1$ and $A_4$ respectively. This system may behave exactly like the system in Figure~\ref{fig:non_linear_prompting} most of the time, but may prove to be more resilient to unexpected circumstances, as each component is more ``intelligent''.

\begin{figure}[!ht]
  \centering
  \includegraphics[width=0.60\columnwidth]{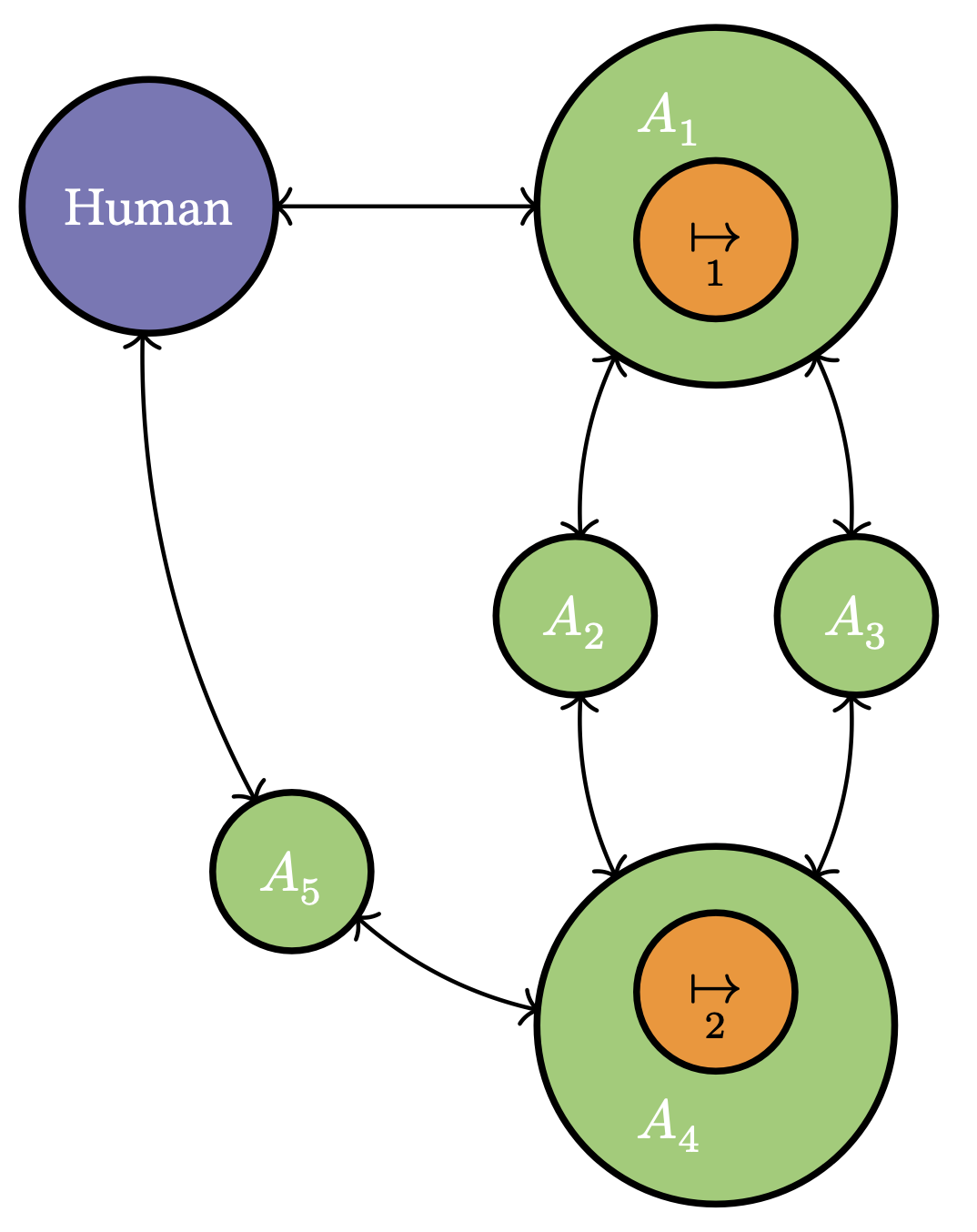}
  \caption{A more realistic projection of the prompting technique from Figure~\ref{fig:non_linear_prompting} as a multi-agent system.}\label{fig:agent_centric_prompting}
\end{figure}
\end{itemize}
As all prompting techniques can be projected to such multi-agent systems, we can conjecture that:

\begin{conjecture}
Results from prompting techniques involving non-linear context can predict similar results from multi-agent systems designed to replicate the same behavior.
\end{conjecture}

This projection or view allows us to generalize all such techniques and apply learnings from one technique to another, and even learnings from multi-agent systems. For example, if new LLM-based multi-agent collaboration research shows that ``A Process Supervising Agent is all you Need'', then we can immediately apply that result to the prompting technique described in BRANCH-SOLVE-MERGE from~\citenp{saha_branch-solve-merge_2023} -- by viewing the ``process supervising agent'' work via a non-linear context lens as illustrated in Figure~\ref{fig:non_linear_prompting}, and then adding the BRANCH-SOLVE-MERGE ``nodes and connections''.

But it does not end there -- because of how flexible natural language is, all non-linear contexts can also be projected to linear contexts. For example, say four agents engage in adversarial interaction as described in~\citenp{xi_rise_2023} (\sect 4.2.2), where they argue about a decision until they reach a consensus. The benefit of this interaction paradigm is that each agent can be instructed to look at the problem from various perspectives. 

This interaction can be elicited within a linear context, where the LLM is prompted with the same decision-making problem but with additional instructions to share a turn-by-turn dialogue where four individuals argue about the decision until they reach a consensus. This has been demonstrated in~\citenp{wang_unleashing_2024}, where a single LLM instance is prompted to produce a transcript of multiple personas (agents) interacting with each other to solve a task. The authors call this ``Solo Performance Prompting''. Their results show that this technique---essentially converting non-linear context (multiple agents collaborating) into linear context (a dialogue transcript)---shows performance gains comparable to those achieved by multi-agent systems on other tasks (\citenp{wang_unleashing_2024} does not directly compare to multi-agent systems).

In \citenp{dong_self-collaboration_2024} the authors describe a similar approach minus the ``dialogue'', where multiple roles (analyst, coder, tester, etc.) are simulated by a single LLM with linear context~\cite{dong_self-collaboration_2024} (§2.2, Eq.~1). The paper shows how this approach outperforms baseline and advanced prompting techniques (such as CoT).

\begin{conjecture}
Performance improvements achieved through multi-agent system architectures can be at least partially replicated using single-LLM prompting techniques\footnote{prompting techniques such as Solo Performance Prompting \cite{wang_unleashing_2024} and Self-Collaboration~\cite{dong_self-collaboration_2024}} that simulate equivalent multi-agent interaction patterns within a linear context.
\end{conjecture}

\subsection{Implications for Synthetic Training Data}\label{sec:synthetic_training_data}

Recent work has demonstrated that synthetic data can effectively enhance model capabilities in various applications, from structured information extraction~\cite{josifoski_exploiting_2023} to visual question answering~\cite{su_sk-vqa_2024}.

A key insight stems from an apparent paradox in LLM systems: while all LLMs are trained on ``linear context'' (sequential text), research and practice show that ``non-linear context'' approaches---such as advanced prompting techniques and multi-agent interactions---are of significant interest and demonstrate superior task performance~\cite{saha_branch-solve-merge_2023,ningSkeletonofThoughtLargeLanguage2023,wei_chain--thought_2023,hong_metagpt_2023,wu_autogen_2023}.

The previous subsection presents an argument for how techniques involving non-linear context can be projected to an equivalent technique utilizing linear context. This can have profound implications when you consider that all LLMs are trained on ``linear context'', i.e., trained on continuous sequences of text. If intermediate steps from advanced prompting techniques like BRANCH-SOLVE-MERGE are projected to linear contexts similar to Solo Performance Prompting~\cite{wang_unleashing_2024} and Self-Collaboration~\cite{dong_self-collaboration_2024} --- then they can also be used as synthetic training data. 

Interestingly, a recent approach called Stream of Search (SoS)~\cite{gandhi_stream_2024} further underscores our perspective on using non-linear or suboptimal reasoning traces for training. SoS demonstrates that when LLMs are trained on branching, backtracking search trajectories---serialized into a linear textual format---they acquire stronger problem-solving capabilities and can even discover new strategies. These findings support Conjecture~3 below, illustrating how self-generated, ``messy'' intermediate steps can serve as valuable synthetic data to improve the performance of an LLM.

\begin{conjecture}
Synthetically generated ``self-collaboration'' transcripts of successful task-solving attempts—whether derived from non-linear prompting techniques or multi-agent collaboration—when used as training data, improve LLM performance in both multi-agent systems and advanced prompting techniques targeting similar tasks.
\end{conjecture}

This idea can be extended further by using existing problems and their real-world deliverables, both intermediate and final, and generating simulated interactions between collaborators as synthetic data. Consider taking the requirements of a completed software project on GitHub, along with pull requests/issue commentary, commit messages, commit diffs in chronological order, and using LLMs to fabricate communication between collaborators -- wouldn't the resulting manuscript, perhaps made to resemble a theater play script, be effective training data?

Thus, our proposed framework of linear and non-linear context (\sectref{sec:linear_nonlinear_context}) along with the agent-centric projection of prompting techniques (\sectref{sec:agent_centric_projection}) presents a lens that could lead to significant advancements in synthetic data generation.

\section{\uppercase{Conclusions}}\label{sec:summary_and_conclusion}

\subsection{Core Arguments}\label{sec:core_arguments}

\begin{itemize}
\item \textsc{Prompt Engineering and Instruction Engineering}: Clearly differentiating the adjustment of the prompt without altering the actual task or the definition of the problem (prompt engineering) and modifying the task to an equivalent task\footnote{one with the same final deliverable} more suitable for LLM systems (instruction engineering) is essential to precise communication and understanding of research in this area.

\item \textsc{Linear and Non-linear Context}: Prompting techniques and resulting task-oriented LLM systems can be classified into having either linear or non-linear context.

\item \textsc{Agent-centric Projection of Prompting Techniques}: We demonstrate approaches that allow prompting techniques with non-linear context to be understood as multi-agent systems and vice versa. This projection provides a framework for analyzing, comparing, and improving both prompting techniques and multi-agent system architectures.
\end{itemize}

\subsection{Implications for Future Research}\label{sec:implications_future}

\begin{itemize}
\item \textsc{Cross-Pollination in Prompting and LLM-based Multi-agent Systems}: The agent-centric projection of prompting techniques may allow us to cross-pollinate research findings in these areas.

\item \textsc{Synthetic Training Data Generation}: Our core arguments suggest two novel approaches for generating high-quality synthetic training data for LLMs: (1) converting successful non-linear prompting traces into linear training data and (2) augmenting real-world task traces with synthetic agent collaboration artifacts. These approaches could provide structured, high-quality data specifically suited for training LLMs in multi-agent and complex reasoning tasks.

\item \textsc{Real-world Applications and Ethical Considerations}: As these systems become more capable, their deployment in real-world scenarios becomes more feasible. With this comes the need for rigorous ethical considerations, especially concerning autonomy, decision making, and human-AI interaction.
\end{itemize}

By establishing the fundamental distinction between linear and non-linear contexts in prompting techniques, and using this to develop an agent-centric projection that reveals deep connections between prompting techniques and multi-agent systems. This framework leads to three key conjectures about the relationship between prompting techniques and multi-agent systems, suggesting that results from one domain can inform the other. Furthermore, we demonstrate how this unified perspective opens up novel approaches to synthetic training data generation, both through the conversion of non-linear prompting traces and through the augmentation of real-world task traces. Our position highlights the untapped potential in viewing prompting techniques through an agent-centric lens, providing concrete directions for improving both the design and training of future LLM systems.

\bibliographystyle{apalike}
{\small
\bibliography{references}}

% \section*{\uppercase{Appendix}}

% If any, the appendix should appear directly after the
% references without numbering, and not on a new page. To do so please use the following command:
% \textit{$\backslash$section*\{APPENDIX\}}

\end{document}